\documentclass{aifrontiers}
\usepackage{aifrontiers}
\usepackage[T1]{fontenc}
\usepackage[utf8]{inputenc}
\usepackage{tabularx,booktabs,array}
\usepackage{caption,wrapfig,needspace}
\usepackage{multicol,enumitem}
\usepackage{inconsolata}
\definecolor{navy}{HTML}{17324D}
\definecolor{teal}{HTML}{137C82}
\definecolor{pale}{HTML}{EAF4F3}

\let\cite\citep
\shorttitle{Persona Simulation for Real-World Decision Making}
\date{\today}
\renewcommand{\AIFLogo}{\raisebox{-0.1ex}{\includegraphics[height=1.5em]{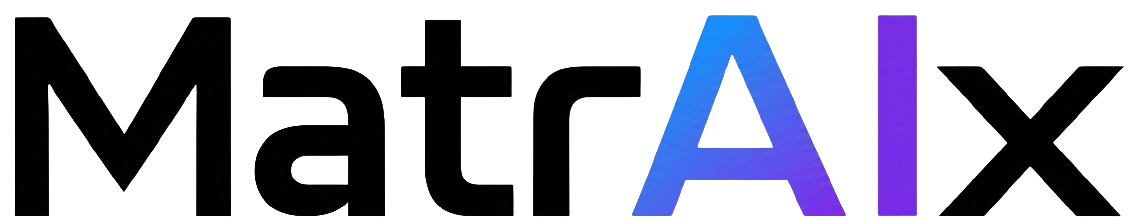}}}

\title{Probe What-If Worlds Before You Ship: \\
Persona Simulation for Real-World Decision Making}

\author{
\normalfont\normalsize
 Julie Zhu\textsuperscript{*}, Jianheng Hou\textsuperscript{*}, Zhixu Silvia Tao\textsuperscript{*}, Shi Bo\textsuperscript{$\dagger$}, Yilan Fan\textsuperscript{$\dagger$}, Yixuan He\textsuperscript{$\dagger$}, Zhiyuan Hu\textsuperscript{$\dagger$}, Shirley Huang\textsuperscript{$\dagger$},
\\ \normalfont\normalsize
  Fangyu Liu\textsuperscript{$\dagger$}, Yijun Wang\textsuperscript{$\dagger$}, Zibu Wei\textsuperscript{$\dagger$}, Haimingyue Xu\textsuperscript{$\dagger$}, Zhen Xu\textsuperscript{$\dagger$}, Zichen Yuan\textsuperscript{$\dagger$}, Jierui Zuo\textsuperscript{$\dagger$}, Jintao Huang,
\\ \normalfont\normalsize
  Yifan Wang, Qianfeng Wen, Jiahan Li, Zhejian Peng, Bowen Yang, Shushu Liang, Yuexing Hao, Xiaomin Li
}

\begin{document}

\begin{abstract}
User research and system testing reveal whether products and services meet user needs, yet recruiting specialized participants, designing controlled what-if comparisons, and repeating studies as designs evolve often bottleneck design iteration and data-driven decision-making. We present a framework and execution infrastructure that translates real-world industry inquiries into controlled experiments with simulated users. Our methodology formally defines the target cohort, experimental stimuli, and decision-relevant metrics, while the pipeline supports large-scale concurrent runs over cohort–stimulus–environment combinations (survey, web, chatbot, and OS app), auditable trial-level logging, cohort-level outcome aggregation, and cross-cohort and cross-variant statistical reports. Across twelve industry partner cases spanning five domains, we show how constructing hard-to-reach or what-if persona cohorts supports exploratory studies and distills simulated responses into testable design hypotheses. Cross-model comparisons on selected cases largely preserve primary conclusions, whereas perturbing or ablating persona attributes alters service outcomes, effect sizes, and inter-group differences, indicating that grounding target cohorts in detailed personas materially shapes simulation results. Together, these studies show that persona-enabled agents can test specified cohorts at design speed, diagnose systems under test, and deliver directional evidence that focuses problem definition, steers design and strategic decisions, and guides further study.

\end{abstract}

\maketitle
\begin{NoHyper}
\begingroup
\renewcommand{\thefootnote}{*}
\footnotetext{Equal contribution.}
\endgroup
\begingroup
\renewcommand{\thefootnote}{\textdagger}
\footnotetext{These authors contributed equally as case-study owners.}
\endgroup
\end{NoHyper}

\section{Introduction}

User testing is integral to product development: it reveals how target users interpret an experience, what hinders adoption, and which design iterations satisfy their needs \cite{gould1985designing,faulkner2003beyond}. Yet live studies face three structural constraints that limit how frequently teams can gather this empirical evidence \cite{dou2025simulatorarena}. First, recruiting participants and running repeated testing cycles becomes prohibitively slow when studies require specialized cohorts defined by specific health circumstances, professional roles, or service requirements. Second, because attributes in live populations are confounded, teams struggle to isolate the effect of a single user attribute by constructing cohorts that match on all other characteristics. Third, exposing existing customers to unproven designs or policies can degrade their experience and damage client relationships during the experiment \cite{kohavi2013online}.

\begin{figure}[t]
\centering
\includegraphics[width=\textwidth]{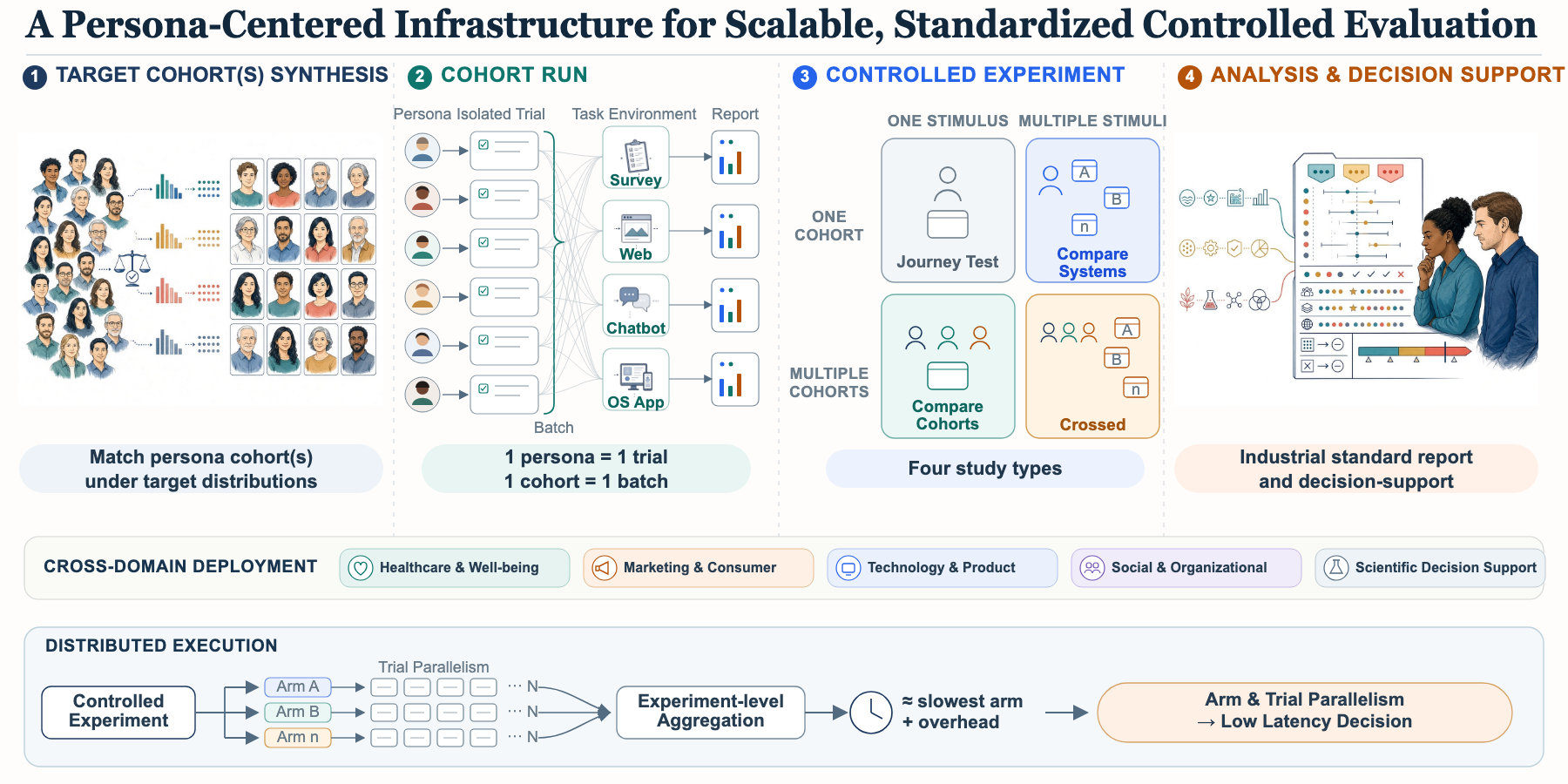}
\caption{The execution pipeline supports scalable, controlled simulated-user evaluation across five application domains.}\label{fig:pipeline}
\end{figure}

Two partner studies illustrate how these constraints arise in practice. In one study, a healthcare provider sought to identify care-access barriers across patients with varying treatment regimens and household resources, where recruiting specialized patient cohorts and disentangling correlated socioeconomic factors are both costly. In another, a business-service partner needed to evaluate how clients would react to revised engagement terms across alternative wordings, without risking live client relationships on untested contract language. Together, hard-to-reach cohorts, confounding across user attributes, and the risk to live traffic motivate conducting controlled what-if comparisons offline before deployment.

Persona-based user simulation removes both obstacles by replacing live cohorts with explicitly specified user profiles \cite{park2022social,bougie2025simuser}: teams can hold background attributes fixed while varying target preferences or behaviors, and repeatedly evaluate evolving design candidates offline without exposing existing customers to untested interfaces. This capability has made simulated users a standard tool for early-stage design exploration \cite{park2022social,park2023generative}, recommendation testing \cite{yoon2024generative,bougie2025simuser}, and assistant evaluation \cite{dou2025simulatorarena}.

Building on an existing persona corpus and execution infrastructure \cite{li2026matraix}, we examine twelve industry partner case studies spanning five application domains: health and well-being, marketing and consumer behavior, technology and product design, social and organizational research, and scientific research. For each partner inquiry, we specify who the target users are, what stimuli they evaluate, and which outcomes govern the decision. These specifications are translated into at-scale controlled experiments with simulated users across four types of environments by our infrastructure.  

\paragraph{Contributions.}
We contribute a common methodology for what-if comparisons of design alternatives and specified persona attributes, covering experimental design, cohort construction, stimulus preparation, and outcome measurement, together with an open-source infrastructure for concurrent, auditable, and repeatable execution; twelve industry partner case studies across four digital environment and five domains that identify participation barriers, audience differences, and design hypotheses; and controlled comparisons examining the sensitivity of selected findings to simulation models and persona information. The case studies establish concrete targets or directions for subsequent human testing with rationale rather than validated predictions of real-user behavior.

\section{Methodology}

Our methodology translates partner inquiries into controlled simulated-user studies. A \emph{persona} is a set of attributes conditioning an LLM's behavior through agent harnesses; a \emph{cohort} is a collection of personas selected for the audience of interest. A \emph{stimulus} is the material or experience evaluated, and an \emph{experiment arm} specifies one cohort--stimulus condition.

\Needspace{.27\textheight}
\subsection{Experimental Design}

\begin{wrapfigure}{r}{.42\textwidth}
\centering
\includegraphics[width=\linewidth]{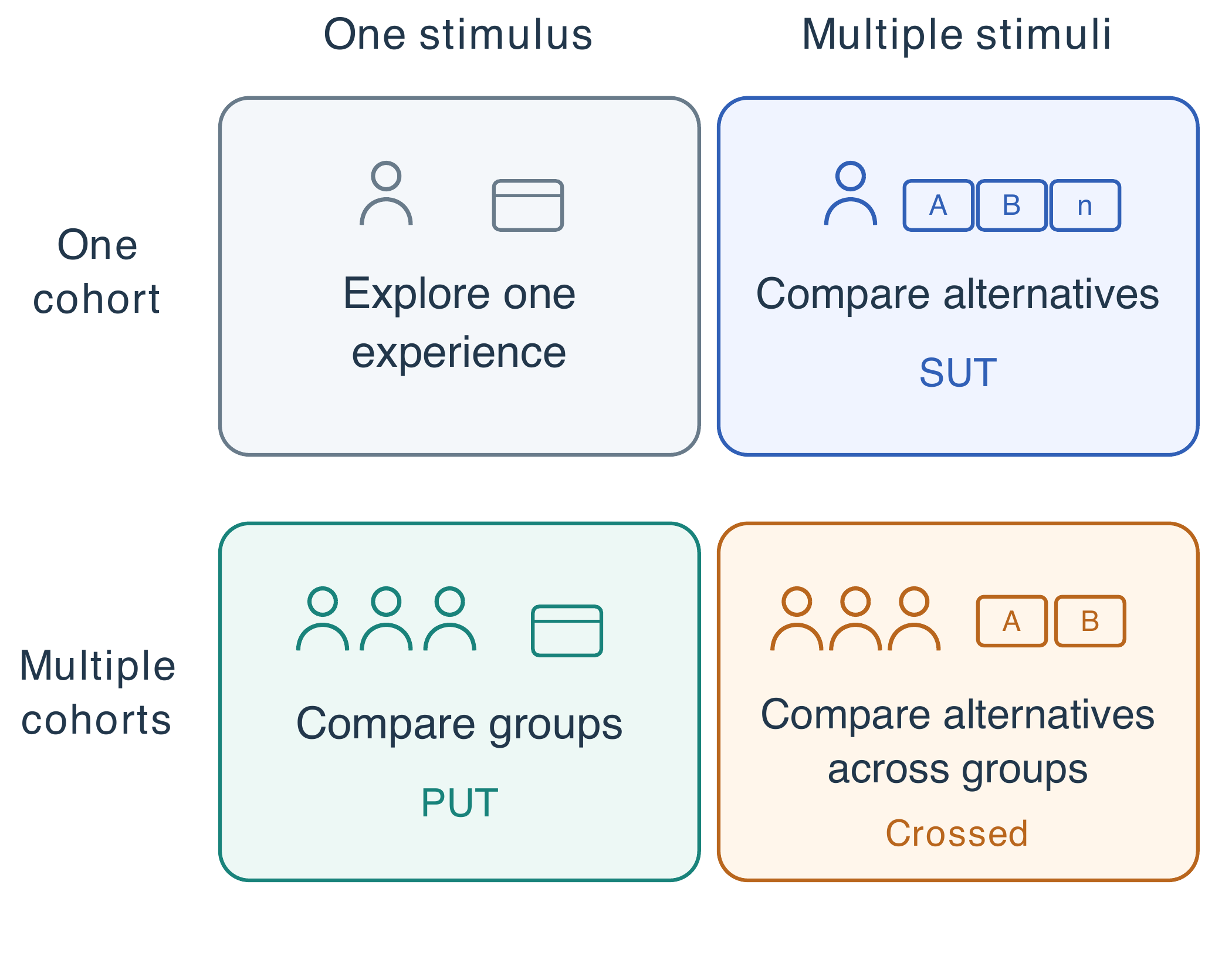}
\caption{Design taxonomy defined by cohorts and stimuli. SUT compares alternatives within one cohort; PUT compares cohorts under one stimulus; crossed designs vary both.}
\label{fig:taxonomy}
\end{wrapfigure}

\paragraph{Define the experiment type.}
We begin with a concrete research question that specifies the target population, the stimulus to be evaluated, the comparison of interest, and the decision the experiment is intended to inform. We organize experiments along two axes for exploring what-if scenarios within the simulation: whether the cohort varies and whether the stimulus varies. A single cohort evaluating a single stimulus constitutes a descriptive evaluation. When the cohort is held fixed and the stimulus varies, the experiment evaluates changes to the product, message, or experience; we refer to this as a \emph{System Under Test} (SUT) experiment. When the stimulus is held fixed and the cohort varies, the experiment compares responses across population segments; we refer to this as a \emph{Persona Under Test} (PUT) experiment. When both vary, a crossed design can be used to estimate stimulus effects across cohorts and examine their interactions. Figure~\ref{fig:taxonomy} summarizes this taxonomy.

\paragraph{Specify cohort attributes.}
We build on prior work that represents personas using 1,290 attribute dimensions whose dependencies are encoded in a directed acyclic graph (DAG)~\citep{li2026matraix}. We first translate the target-population description into constraints on these dimensions. When an exact match exists, we map the description directly to the corresponding attributes; for example, ``people aged 18--34'' maps to the \texttt{18--24} and \texttt{25--34} age brackets. When the target construct is not represented directly, we approximate it using one or more related attributes. For example, high music engagement can be approximated by listening to music ``daily'' or ``all day,'' although listening frequency is not equivalent to engagement. We make such limitations explicit and check the resulting constraints against the dependencies encoded by the DAG. Logically inconsistent combinations---for example, simultaneously specifying that a persona lives alone and belongs to a multi-person household---are excluded.

\paragraph{Sample the cohort.}
After defining the relevant attributes, we estimate the cohort size and select sampling strategy according to the experiment's purpose. We consider five cohort-composition strategies. First, \emph{reference-data-informed sampling} sets target proportions using relevant human data; for example, a healthcare study may use population and clinical-study data to specify age, body mass index (BMI), sex, and race or ethnicity distributions. Second, \emph{balanced comparison} assigns equal or similar representation to the groups being compared; for example, a study of an AI research assistant may include equal numbers of senior researchers, wet-lab trainees, computational biologists, and molecular geneticists. Third, \emph{targeted oversampling} increases the representation of an important but uncommon group; for example, a study of a service for remotely inspecting vehicles, properties, suppliers, or documents may oversample personas with relevant transaction experience. Fourth, \emph{broad coverage} distributes personas across diverse backgrounds and circumstances without attempting to reproduce population proportions; for example, a professional-networking study may include varied demographic backgrounds, professional roles, and levels of digital experience. Finally, \emph{stress-test sampling} increases the representation of personas facing task-relevant challenges; for example, a community-health study may emphasize affordability constraints, limited access to care, and difficulty understanding health information. These strategies may be adapted or combined when required by the research question. When an experiment requires multiple cohorts, each cohort is constructed using the same procedure. Balanced, exploratory, and stress-test cohorts do not estimate population proportions.

\paragraph{Prepare the stimulus and task.}
The stimulus may be an existing product, stakeholder-provided material, or an artifact constructed for the experiment. It is preserved in the form that a real user would encounter whenever possible. For controlled comparisons, only the intended experimental factor is varied. For example, a message-framing experiment may change the framing while keeping the factual content, questionnaire, question order, identifiers, and response scales constant.

Each experiment also includes a task specification describing the situation, the action expected from the simulated user, and, for interactive tasks, the available actions and stopping conditions. Instructions are written neutrally and include only information that would reasonably be available to the target user.

\paragraph{Define outcome metrics.}
Studies specify decision-relevant outcomes, diagnostic measures, and guardrails. Numerical ratings, categorical choices, and explanations retain their scales, category definitions, and valid-response denominators.

\subsection{Execution Infrastructure for Large-Scale Experiments}
\label{sec:infrastructure}

A trial comprises one simulated persona completing a task under one arm. Each persona receives only the stimulus assigned to that arm, preserving the intended comparison. The pipeline fixes cohort membership, sampling seeds, stimulus versions, instructions, and model configurations, and executes independent trials concurrently across workers. Specifications can be reused after changing a stimulus, design, or simulation model, allowing follow-up studies to retain the original cohort and outcome definitions. Controlled model comparisons aim to retain cohort, stimulus, task, and metrics; actual endpoint and prompt-assembly differences are reported because they limit model-only or persona-only interpretations. Throughput depends on interaction format, model latency, service quotas, and computing resources.

Each trial retains the persona, rendered inputs, response or interaction trajectory, and structured outcome. Aggregation preserves links to these records and reports failed or missing trials, allowing cohort-, arm-, and subgroup-level findings to be audited.

\paragraph{Statistical reporting.}
Reports preserve the valid-response denominator for each outcome and summarize responses by experiment arm and subgroup. For numerical comparisons, the pipeline provides effect estimates and uncertainty intervals while retaining links to the underlying trials. The analysis unit and any shared-persona or group structure are specified alongside each comparison. The statistical procedures depend on the comparison design; Appendix~\ref{app:stats} details the statistical procedures and their interpretation.

\section{Twelve Cross-Domain Industry Case Studies}\label{sec:applications}

\begin{table}[t]
\centering
\caption{Twelve industry partner case studies spanning five application domains and their principal simulated-user findings. Case identifiers link to the corresponding appendix reports.}
\label{tab:fields}
\small
\begin{tabularx}{\textwidth}{@{}>{\raggedright\arraybackslash}p{.14\textwidth}>{\raggedright\arraybackslash}p{.24\textwidth}>{\raggedright\arraybackslash}p{.11\textwidth}>{\raggedright\arraybackslash}X@{}}
\toprule
\textbf{Application domains} & \textbf{Cases} & \textbf{Test environment} & \textbf{Principal finding} \\
\midrule
Health \& well-being
& \hyperref[app:music-wellbeing]{C1: Music-wellbeing onboarding} & Survey
& Asking how users felt built more trust than emphasizing science or adopting a peer-companion tone. \\
& \hyperref[app:healthcare-access]{C10: Healthcare access barriers} & Survey
& Cost barriers were widespread, but rural users faced greater travel barriers and adults aged 31--50 reported greater time pressure. \\
\addlinespace
Marketing \& consumer behavior
& \hyperref[app:subscription-pricing]{C3: Subscription pricing and renewal} & Survey
& A cheaper introductory offer attracted interest, but recurring payments discouraged subscription. \\
& \hyperref[app:remote-service]{C5: Remote-service marketplace} & Survey
& Users were more concerned about trusting the person performing the service than about its price. \\
& \hyperref[app:engagement-letter]{C9: Engagement-letter evaluation} & Survey
& Revising payment and liability terms improved acceptance, but distrust remained a barrier to signing. \\
\addlinespace
Technology \& product design
& \hyperref[app:professional-network]{C6: Professional-network signup} & Web
& Users understood how to sign up but hesitated over identity verification and document privacy. \\
& \hyperref[app:industrial-website]{C8: Industrial website engagement} & Web
& Visitors were more willing to explore than contact sales; ``try for free'' links led to sales forms that did not meet their expectations. \\
& \hyperref[app:home-assistant]{C11: Home assistant} & Chatbot/OS
& Comfort-first preferences were satisfied more often than shorter-runtime preferences across three simulation models. \\
& \hyperref[app:car-assistant]{C12: Car assistant} & Chatbot/OS
& Proactive preference disclosure improved service when canceling a route stop; a breakfast-stop task showed model-dependent differences. \\
\addlinespace
Social \& organizational research
& \hyperref[app:workplace-assessment]{C7: Workplace survey responses} & Survey
& Workers gave more favorable answers in supportive workplaces; changing models altered responses more than repeating the same model. \\
\addlinespace
Scientific research
& \hyperref[app:clinical-trial]{C2: Clinical-trial screening review} & Survey
& Pharmacists prioritized medication compatibility, while molecular specialists prioritized genetic confirmation in the same screening review. \\
& \hyperref[app:research-priority]{C4: Scientific research prioritization} & Survey
& Scientists chose different research priorities even when their comprehension scores showed no difference. \\
\bottomrule
\end{tabularx}

\end{table}

Table~\ref{tab:fields} summarizes twelve studies across five domains; protocols and results appear in the Appendix. The cases illustrate two uses: constructing repeatable studies with hard-to-reach or matched what-if cohorts and deriving testable design hypotheses from simulated responses.

\subsection{Constructing Hard-to-Reach and What-If Cohorts}

The methodology supports both constructing cohorts for specialized audiences and creating matched persona cohorts that differ only in specified attributes. To reach specialized populations, the healthcare-access study (\hyperref[app:healthcare-access]{C10}) evaluates 900 personas across nine region--age strata in India with specified clinical conditions, treatment arrangements, and household resources, mapping care-access barriers across intersecting socioeconomic circumstances. Similarly, the clinical-trial screening study (\hyperref[app:clinical-trial]{C2}) assigns 200 personas from each of four specialist roles---including oncology pharmacists and molecular pathologists---to review the same patient record, isolating role-specific evaluation criteria without repeatedly recruiting scarce clinicians.

The car-assistant study (\hyperref[app:car-assistant]{C12}) illustrates controlled what-if cohort construction: each of 50 base personas is paired with a counterfactual version that holds background attributes and travel priorities fixed while varying whether the user discloses preferences proactively or only when prompted. In the stop-cancellation task, proactive disclosure increases preference satisfaction by 18--44 percentage points across all three simulation models, pinpointing active preference elicitation as a high-priority mechanism for live validation. Across all three studies, simulated cohorts serve to generate targeted hypotheses for human testing rather than to replace empirical evidence from live populations.

\subsection{From Simulated Responses to Design Implications}

Linking aggregate outcomes to individual choices, explanations, and interaction logs helps distinguish task difficulty from low perceived value or unwillingness to participate.

Professional-network signup (\hyperref[app:professional-network]{C6}) asked whether prospective users trusted a networking website enough to join. About 98\% reached the signup entrance, yet only about 1\% stated that they would register. The website was generally rated as clear, while explanations identified concerns about identity verification and document retention. Improving navigation alone would therefore leave the reported concerns unresolved; the study directed attention to privacy disclosures and verification procedures.

The subscription pricing and renewal study (\hyperref[app:subscription-pricing]{C3}) exposed a similar gap between perceived value and willingness to join. About 98\% considered a cheaper introductory offer a good deal, but only about 51\% would subscribe. Non-subscribers identified the recurring commitment as the main barrier. The reported hesitation concerned the ongoing payment obligation, directing the next test toward renewal terms.

Persona attributes also helped interpret inter-group differences in priorities. Scientific research prioritization (\hyperref[app:research-priority]{C4}) asked which unresolved immunology questions four professional groups would investigate next. Their choices differed despite no observed difference on the comprehension measure, with molecular geneticists showing the largest departure from senior investigators. A single pooled preference would conceal this disciplinary variation; separate group analyses identified perspectives to include in expert consultation.

These explanations identify hypotheses for product revision and subsequent human studies; they do not establish that the proposed changes improve real-user outcomes.


\section{Robustness and Ablation Analysis}

We examine whether selected findings recur across user-simulation models and how they change when persona attributes are varied or descriptions omitted.

\subsection{Cross-Model Robustness}\label{sec:cross-model-robustness}

\begin{figure}[!htbp]
\centering
\begin{minipage}[t]{.485\linewidth}
\vspace{0pt}
\centering
\includegraphics[width=\linewidth]{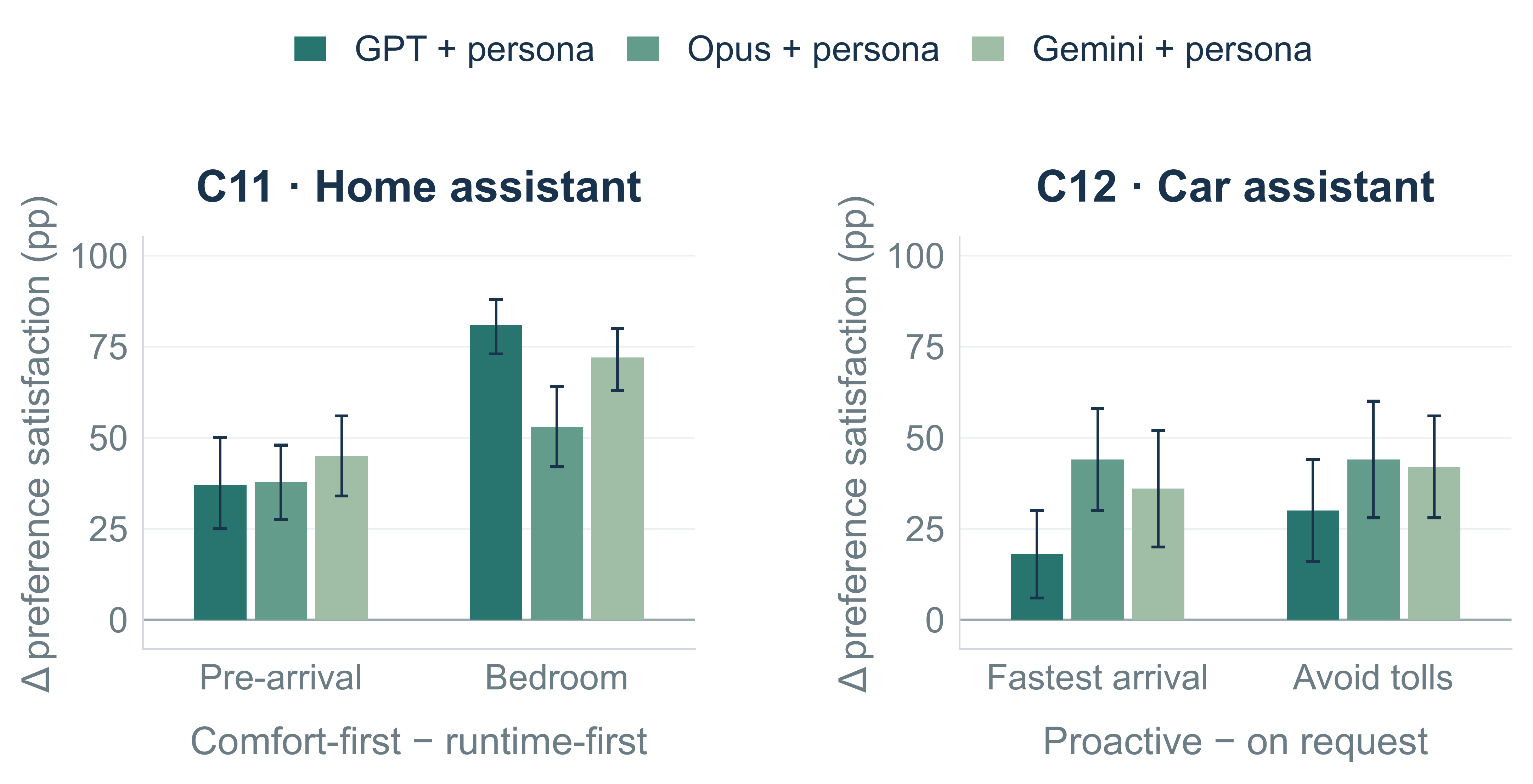}
\caption{Persona changes affect assistant preference satisfaction. C11: preparing the home or maintaining bedroom comfort, comparing comfort-first versus runtime-first personas (the latter prefer shorter device runtime). C12: canceling a route stop, comparing proactive versus on-request disclosure of travel preferences. Bars show first-minus-second satisfaction rates (percentage points); colors identify user-simulation models. Error bars show paired-bootstrap 95\% confidence intervals.}
\label{fig:model-persona-comparisons}
\end{minipage}%
\hfill
\begin{minipage}[t]{.485\linewidth}
\vspace{0pt}
\centering
\includegraphics[width=\linewidth]{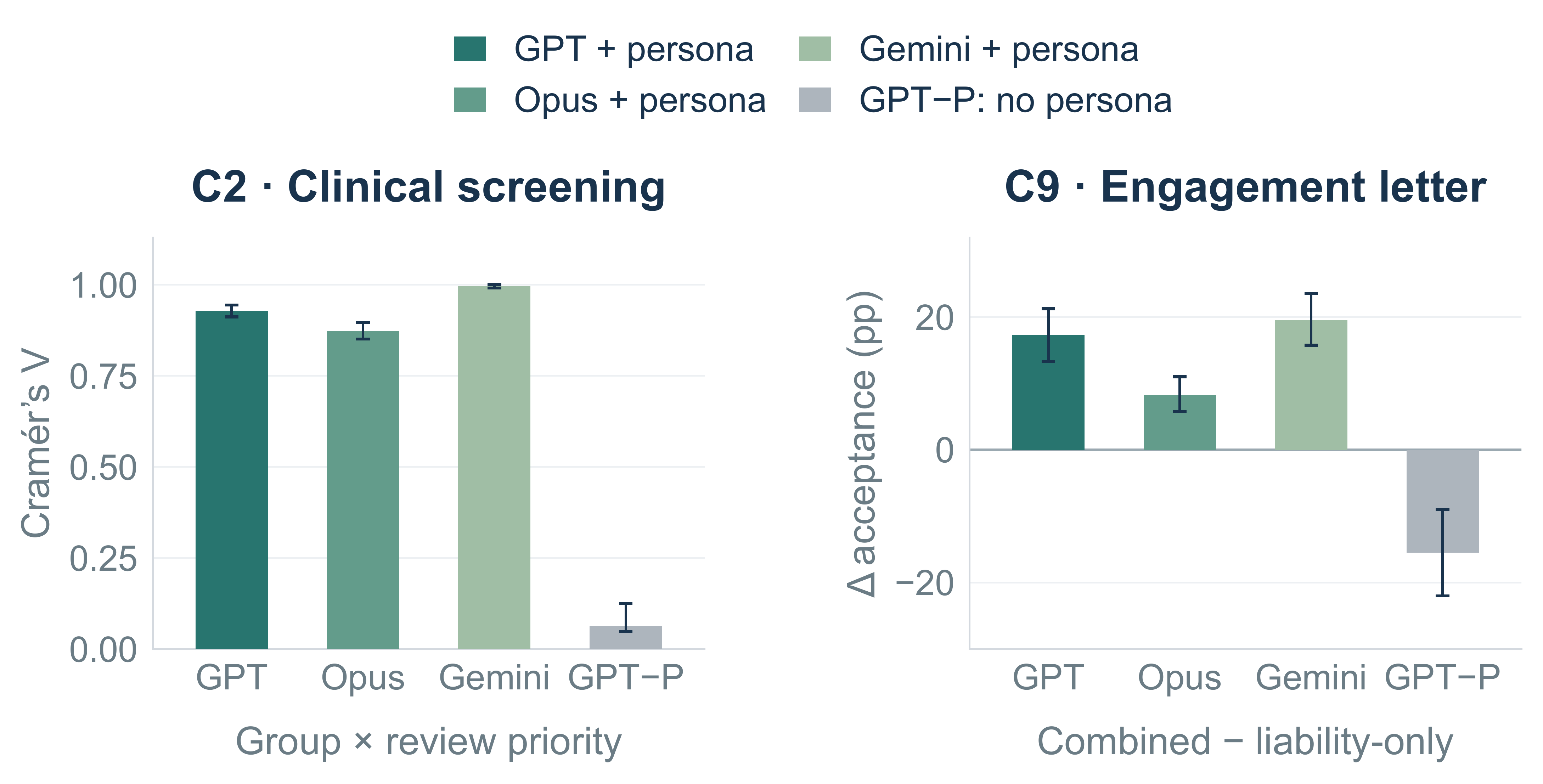}
\caption{Persona information shapes simulated judgments. C2: association between professional group and clinical-screening review priority (Cram\'{e}r's $V$; higher means stronger association). C9: client acceptance of combined payment-and-liability revisions minus liability-only revisions to an engagement letter (percentage points). Green bars use personas; gray denotes GPT without personas. Error bars show bootstrap 95\% confidence intervals.}
\label{fig:persona-information-comparisons}
\end{minipage}
\end{figure}

Figures~\ref{fig:model-persona-comparisons} and~\ref{fig:persona-information-comparisons} compare GPT-5.6-Sol, Claude Opus~5, and Gemini 3.7 Flash, distinguishing recurring directions from agreement in effect magnitude.

All three models yield higher preference satisfaction for comfort-first than shorter-runtime personas in both home tasks. For canceling a car route stop, proactive disclosure outperforms on-request disclosure under both travel priorities (Figure~\ref{fig:model-persona-comparisons}). Home contrasts average over disclosure conditions within matched personas, whereas car contrasts hold travel priority fixed while varying disclosure. These selected contrasts recur across the evaluated configurations; the breakfast-stop task instead shows model-dependent effects (Appendix~\ref{app:car-assistant}).

Professional group remains strongly associated with clinical-screening review priority, and all three persona models favor combined payment-and-liability revisions over liability-only revisions (Figure~\ref{fig:persona-information-comparisons}). This supports shared hypotheses for follow-up testing, not interchangeable quantitative estimates. Home/car assistant endpoints and engagement prompt assembly differ across configurations, limiting model-only interpretations.

\subsection{Effects of Persona Conditioning}

\paragraph{Perturbing persona attributes.}
As shown in Figure~\ref{fig:model-persona-comparisons}, controlled persona-attribute perturbations provide what-if comparisons of support across simulated user segments: the home assistant under-serves runtime-minimizing users (\hyperref[app:home-assistant]{C11}, Table~\ref{tab:home-assistant-results}), while the car assistant under-serves users who disclose preferences only when asked (\hyperref[app:car-assistant]{C12}, Table~\ref{tab:car-assistant-results}). These gaps motivate tests of runtime-aware planning and preference clarification. Trust also strongly differentiated engagement-letter acceptance (Appendix~\ref{app:engagement-letter}).

\paragraph{Ablating persona information.}
As a separate sensitivity check, omitting persona specifications substantially weakens inter-group differences and can even reverse simulated design rankings (Figure~\ref{fig:persona-information-comparisons}). In clinical-trial screening, the association between professional group and review priority collapses from Cram\'{e}r's $V=.928$ to $.063$ without persona descriptions (Appendix~\ref{app:clinical-trial}). In engagement-letter evaluation, the combined-minus-liability-only acceptance difference changes from $+17.25$ to $-15.50$ percentage points, reversing the simulated ranking of the revisions (Appendix~\ref{app:engagement-letter}). Different prompt assembly prevents attributing this reversal solely to persona removal, but the comparison shows that audience specification can materially affect which design the simulation favors.

\section{Discussion and Conclusion}

Across twelve industry partner studies, simulated feedback turned broad evaluation questions into specific design choices. It distinguished interest from willingness to participate, revealed professional priorities that pooled results would conceal, and exposed gaps in assistants' support for different user preferences and disclosure styles. Model comparisons supported shared directions for pilot testing, while persona ablations showed that audience information can change which revision is selected.

The contribution is a common methodology and infrastructure for making these comparisons repeatable and statistically traceable. By connecting specified audiences to controlled what-if comparisons of designs and persona attributes, the framework helps prioritize candidates for live testing and gives subsequent user studies a sharper focus, without establishing causal effects in real users.

\section*{Limitations}
Incomplete human reference data and imperfect behavioral representation limit the precision of real-world estimates from these studies. The reported intervals quantify variation within the specified simulation designs (Appendix~\ref{app:stats}); they do not measure discrepancies between simulated and real-user behavior.

\section*{Ethical Considerations}
All reported user responses are synthetic. Partner materials and case-level records remain confidential, while the public repository provides the study infrastructure. Disclosure of case descriptions and aggregate findings follows case-specific partner authorization.

\bibliographystyle{plainnat}
\bibliography{reference}

\clearpage
\section*{Authors and Affiliations}
\label{sec:author-affiliations}

\begingroup
\raggedright
\normalfont\normalsize
\noindent \mbox{Julie Zhu\textsuperscript{1,*}}, \mbox{Jianheng Hou\textsuperscript{8,*}}, \mbox{Zhixu Silvia Tao\textsuperscript{2,*}}, \mbox{Shi Bo\textsuperscript{7,$\dagger$}}, \mbox{Yilan Fan\textsuperscript{4,$\dagger$}}, \mbox{Yixuan He\textsuperscript{3,$\dagger$}}, \mbox{Zhiyuan Hu\textsuperscript{1,$\dagger$}}, \mbox{Shirley Huang\textsuperscript{6,$\dagger$}}, \mbox{Fangyu Liu\textsuperscript{8,$\dagger$}}, \mbox{Yijun Wang\textsuperscript{6,$\dagger$}}, \mbox{Zibu Wei\textsuperscript{5,$\dagger$}}, \mbox{Haimingyue Xu\textsuperscript{16,$\dagger$}}, \mbox{Zhen Xu\textsuperscript{10,$\dagger$}}, \mbox{Zichen Yuan\textsuperscript{9,$\dagger$}}, \mbox{Jierui Zuo\textsuperscript{17,$\dagger$}}, \mbox{Jintao Huang\textsuperscript{11}}, \mbox{Yifan Wang\textsuperscript{12}}, \mbox{Qianfeng Wen\textsuperscript{13}}, \mbox{Jiahan Li\textsuperscript{4}}, \mbox{Zhejian Peng\textsuperscript{14}}, \mbox{Bowen Yang\textsuperscript{10}}, \mbox{Shushu Liang\textsuperscript{6}}, \mbox{Yuexing Hao\textsuperscript{15,1}}, \mbox{Xiaomin Li\textsuperscript{15,6}}.

\subsection*{Affiliations}
\small
\begin{multicols}{2}
\begin{enumerate}[leftmargin=1.6em,itemsep=0.15em,topsep=0pt]
  \item Massachusetts Institute of Technology
  \item Princeton University
  \item Arizona State University
  \item New York University
  \item University of California, Los Angeles
  \item Harvard University
  \item Boston University
  \item University of Southern California
  \item Washington University in St. Louis
  \item Columbia University
  \item The Ohio State University
  \item Purdue University
  \item University of Toronto
  \item University of Illinois Urbana-Champaign
  \item MatrAIx
  \item VEGA
  \item University of Washington
\end{enumerate}
\end{multicols}

\noindent\textsuperscript{*}Equal contribution.\par
\smallskip
\noindent\textsuperscript{$\dagger$}These authors contributed equally as case-study owners.\par
\medskip

\endgroup

\section*{Acknowledgement}

We thank Kevin Zhou for contributing model credits that supported the experiments in this study.

\clearpage

\appendix
\section{Extended Related Work}
\label{app:related}
\label{sec:related}

\paragraph{Simulated users for evaluation.}
Social Simulacra introduced generated interactions for early design exploration \cite{park2022social}, while Generative Agents demonstrated interactive behavior in shared environments \cite{park2023generative}. Subsequent studies developed simulated users for conversational recommendation \cite{yoon2024generative,meshi2026convapparel}, recommender evaluation \cite{bougie2025simuser}, and assistant benchmarking \cite{dou2025simulatorarena}. PersonaEval connects persona-based users to surveys, chatbots, and web applications while recording interaction trajectories and outcomes \cite{liu2026personaeval}. AgentPersonaBench benchmarks adherence to specified persona attributes during task execution \cite{huang2026agentpersonabenchbenchmarkingpersonadrivenuser}. Building on these evaluation capabilities, we develop a common study methodology that connects industry partners' evaluation needs to audience specification, controlled comparisons, and statistical reporting. Twelve partner studies across five application fields show how this methodology informs concrete design decisions.

\paragraph{Audience specification and behavioral grounding.}
Turing experiments and silicon-sample studies demonstrate that LLMs can reproduce selected behavioral patterns \cite{aher2023turing,argyle2023many}. Population and survey audits further show that representation depends on the audience, task, and measurement \cite{santurkar2023opinions,bisbee2024replacements,wang2025sociobench,hu2026simbench,peng2026funhouse}. Human-grounded agents and matched-response datasets provide references for evaluating this dependence \cite{park2026grounded,toubia2025twin}. Human comparisons and behavioral evaluations identify task-dependent differences in simulated behavior \cite{zhou2026sim2real,zhu2026realusersim,chen2026omni,du2026twinvoice,pohl2026belief,chen2026fail}. Our studies examine how audience specification shapes the resulting insights, using persona ablations and model comparisons to distinguish findings that persist from those that depend on the simulation configuration.

\paragraph{Simulation infrastructure and application studies.}
Concordia, AgentSociety, and OASIS provide reusable environments for social simulation \cite{vezhnevets2023concordia,piao2025agentsociety,yang2024oasis}. MicroVerse provides an instrument for measuring identity drift in long-horizon multi-agent simulations \cite{ng2026microverse}. Methodological syntheses emphasize explicit design and validation choices \cite{buck2026cookbook,larooij2026validation}. Building on prior infrastructure \cite{li2026matraix}, we connect cohort construction, isolated trial execution, and outcome reporting through a common study specification, allowing partner evaluations to be repeated and inspected across application settings.

\paragraph{Statistical interpretation.}
Uncertainty quantification and human-data-assisted inference provide methods for analyzing synthetic responses \cite{huang2025uq,krsteski2026valid}. Recent work formalizes the assumptions required to connect simulated responses and ratings to human outcomes \cite{hullman2026behavioral,persson2026surrogacy,guerdan2026doubly}. Our reporting protocol links each estimate and uncertainty interval to its comparison design and analysis unit, with uncertainty interpreted within the simulation design. This supports consistent interpretation across partner studies.

\section{Case Reports}\label{app:case-report}
\label{sec:appendix}
\subsection{C1: Music-Wellbeing Onboarding}\label{app:music-wellbeing}

\paragraph{Summary.}
A music-wellbeing provider asked which opening message would encourage young users to begin a personalized journey while preserving trust, comfort, and privacy confidence.

\paragraph{Cohort.}
The persona-conditioned studies targeted a 1,000-person young-user cohort. GPT-5.6-Sol and Opus~5 reports identify the same cohort and seed~42. A separate GPT-5.6-Sol baseline omitted individual persona descriptions.

\paragraph{Task and design.}
Each study compared science-informed, warm peer-companion, and check-in-first messages in isolated trials, with 1,000 simulation trials per framing in each study. Primary outcomes were trust, intent to begin, and anticipated journey choice; comfort and privacy confidence served as guardrails. Table~\ref{tab:music-wellbeing-results} reports score differences calculated from the local trial records on a 1--7 scale.

\paragraph{Results.}
GPT and Opus both favored check-in-first over science-informed framing, with trust gains of 0.252 and 0.230 points and intent gains of 0.272 and 0.151 points. Warm peer-companion framing reduced both outcomes. Without personas, the check-in-first trust gain was 0.035 points, while intent increased by 0.351 points. The warm-peer trust penalty increased from 0.093 to 0.640 points without personas.

\par\medskip\noindent
\begin{minipage}{\columnwidth}
\small
\captionsetup{hypcap=false}
\captionof{table}{Music-wellbeing score differences from science-informed framing.}
\label{tab:music-wellbeing-results}
\begin{tabularx}{\columnwidth}{@{}Xrrr@{}}
\toprule
\textbf{Model / condition} & \textbf{Trust} & \textbf{Intent} & \textbf{Privacy} \\
\midrule
\multicolumn{4}{@{}l}{\textit{Check-in first}} \\
Opus 5 & $+0.230$ & $+0.151$ & $+0.861$ \\
GPT-5.6-Sol & $+0.252$ & $+0.272$ & $+0.488$ \\
GPT, no persona & $+0.035$ & $+0.351$ & $+0.722$ \\
\addlinespace
\multicolumn{4}{@{}l}{\textit{Warm peer companion}} \\
Opus 5 & $-0.208$ & $-0.307$ & $-0.161$ \\
GPT-5.6-Sol & $-0.093$ & $-0.077$ & $-0.289$ \\
GPT, no persona & $-0.640$ & $-0.599$ & $-0.209$ \\
\bottomrule
\end{tabularx}
\par\smallskip{\footnotesize\textit{Note.} Entries are local-data mean contrasts on a 1--7 scale. Valid science/warm/check-in counts are 957/960/954 (Opus), 999/1,000/1,000 (GPT), and 1,000/1,000/1,000 (no persona). Paired contrasts use 916 check-in and 920 warm-peer complete pairs for Opus, and 999 for GPT; no-persona arms are independent. The trust contrasts and their intervals are reported here and in Appendix~\ref{app:stats}.}
\end{minipage}
\par\medskip

\paragraph{Insights.}
Agreement between the two persona-conditioned models motivates pilot testing of optional check-ins rather than assuming that warmer language builds trust. The no-persona baseline shows a different response pattern: check-in-first trust benefits weaken without profiles, while intent benefits persist. The directly estimated difference in check-in-first trust effects was +0.217 points (95\% CI [0.183, 0.251]). The results concern simulated messaging responses rather than observed uptake or therapeutic efficacy.


\subsection{C2: Clinical-Trial Screening Review}\label{app:clinical-trial}

\paragraph{Summary.}
This study asked how professional reviewer groups evaluate the same clinical-trial screening assistant and prioritize evidence when its eligibility assessment overlooks unresolved requirements.

\paragraph{Cohort.}
The updated four-group study uses 800 personas, with 200 in each study-assigned group: oncology pharmacists, molecular genetics or pathology specialists, clinical research coordinators, and clinical investigators. The retired trainee group is excluded from this revision. Study-specific attributes specify professional responsibilities and review dimensions.

\paragraph{Task and design.}
All groups reviewed one fictional postoperative ADAURA screening record against supplied historical rules, forming a PUT comparison. An assistant marked eligibility satisfied despite pending central EGFR confirmation and current rifampin use conflicting with an exclusion. GPT-5.6-Sol, Claude Opus~5, and Gemini 3.7 Flash each evaluated the 800-persona cohort. A GPT no-persona condition retained the task and bookkeeping IDs but omitted persona information. Each configuration yielded 800 effective responses. The primary outcome was review priority; eligibility judgment and blocking-issue identification were secondary checks.

\paragraph{Results.}
All four professional groups shared their modal review priority across the three persona models (Table~\ref{tab:clinical-trial-results}). Pharmacists prioritized medication compatibility, molecular/pathology specialists prioritized molecular evidence, and coordinators and investigators prioritized complete eligibility checking.

\par\medskip\noindent
\begin{minipage}{\columnwidth}
\small
\captionsetup{hypcap=false}
\captionof{table}{Modal screening-review priorities across three persona models.}
\label{tab:clinical-trial-cohort}
\label{tab:clinical-trial-results}
\begin{tabularx}{\columnwidth}{@{}Xrrr@{}}
\toprule
\textbf{Reviewer group} & \textbf{GPT} & \textbf{Opus} & \textbf{Gemini} \\
\midrule
Oncology pharmacy & B: 99.5 & B: 97.5 & B: 100 \\
Molecular / pathology & A: 100 & A: 66.0 & A: 100 \\
Research coordination & C: 97.0 & C: 100 & C: 100 \\
Clinical investigation & C: 80.5 & C: 100 & C: 99.0 \\
\bottomrule
\end{tabularx}
\par\smallskip{\footnotesize\textit{Note.} Modal option and percentage; $n=200$ per group and model. A: molecular evidence; B: medication compatibility; C: complete, documented eligibility checking. Results use the updated four-group aggregation (October 11, 2026), including one GPT and nine no-persona supplemental runs. Figure~\ref{fig:persona-information-comparisons} reports group--priority association with pointwise intervals.}
\end{minipage}
\par\medskip

\paragraph{Persona ablation.}
Group--priority association weakened from Cram\'{e}r's $V=.928$ with GPT personas to $.063$ without them. The no-persona control favored complete eligibility checking in all four bookkeeping groups (58.0--63.0\%), whereas persona-conditioned modes differed by professional role. These descriptive estimates replace the earlier five-group analysis; its contrast interval is not reused for the updated data.

\paragraph{Insights.}
Professional context differentiated review priorities, and all three persona models preserved the role-specific modal pattern despite differences in concentration. These results motivate role-specific evidence presentation; the study-assigned review dimensions and absence of matched human responses limit claims about real professional preferences.

\subsection{C3: Subscription Pricing and Renewal}\label{app:subscription-pricing}

\paragraph{Summary.}
A subscription-box provider sought to understand how introductory pricing affected willingness to join and whether initial interest extended to recurring payments.

\paragraph{Cohort.}
The same 500-persona cohort represented English-speaking women aged 18--54 in North America, spanning income levels, life stages, and economic motivations.

\paragraph{Task and design.}
Using GPT-5.6-Luna, personas evaluated five introductory prices: \$9.99, \$29.99, \$39.99, \$49.99, and \$59.99. Product selection, membership benefits, and the \$79.99 quarterly renewal price remained fixed; advertised discounts changed with price. Each condition produced 500 completed evaluations, totaling 2,500 trials.

\paragraph{Results.}
Table~\ref{tab:subscription-pricing-results} shows similar subscription rates at \$29.99 and \$39.99, with substantially lower acceptance at the two higher prices. At \$39.99, 489/500 rated the offer a good deal, but only 257 would subscribe. Across all conditions, no persona affirmed that quarterly renewal was acceptable; responses were unsure or negative. Every non-subscriber identified recurring commitment as the main barrier.

\par\medskip\noindent
\begin{minipage}{\columnwidth}
\small
\captionsetup{hypcap=false}
\captionof{table}{Stated subscription acceptance across introductory prices.}
\label{tab:subscription-pricing-results}
\begin{tabularx}{\columnwidth}{@{}lrr@{}}
\toprule
\textbf{Introductory price} & \textbf{Yes / valid $N$} & \textbf{Rate} \\
\midrule
\$9.99 & 399/500 & 79.8\% \\
\$29.99 & 251/500 & 50.2\% \\
\$39.99 & 257/500 & 51.4\% \\
\$49.99 & 84/500 & 16.8\% \\
\$59.99 & 78/500 & 15.6\% \\
\bottomrule
\end{tabularx}
\par\smallskip{\footnotesize\textit{Note.} The same cohort evaluates every price; the quarterly renewal price remains \$79.99. Entries are descriptive simulated choices, without inferential tests or intervals.}
\end{minipage}
\par\medskip

\paragraph{Insights.}
Introductory-price approval did not imply subscription acceptance or confidence in renewal. The findings motivate separate tests of renewal terms and retention. They describe simulated stated preferences rather than observed purchases or an established optimal price.

\subsection{C4: Scientific Research Prioritization}\label{app:research-priority}

\paragraph{Summary.}
This study examined whether simulated scientists with different professional backgrounds prioritized different unresolved questions in a TCF7/CD8 research-planning task.

\paragraph{Cohort.}
Four synthesized cohorts each comprised 250 personas: senior tumor-immunology principal investigators (PIs) or research scientists, wet-lab doctoral or postdoctoral researchers, computational biologists, and molecular geneticists. Each group contributed 25\% of the 1,000-persona study.

\paragraph{Task and design.}
Using GPT-5.6-Luna, all cohorts evaluated the same research-choice task, forming a PUT comparison with senior investigators as the reference cohort. The questionnaire included primary and secondary categorical choices, labeled B1 and B2, an evidence-comprehension check, and a complete-valid-response indicator. All 1,000 trials completed. The report specified 95\% confidence and 12 comparisons without multiple-testing correction.

\paragraph{Results.}
Both choice distributions differed from the reference in every other cohort (all six $p<.001$; Table~\ref{tab:research-priority-audit}). Molecular geneticists showed the largest differences, with Cram\'{e}r's $V=0.46$ for B1 and $0.45$ for B2, compared with $0.25$--$0.28$ for the other groups. Evidence-comprehension contrasts were zero percentage points ($p=1.000$). Complete-valid-response rates were 0.4 percentage points lower in each nonreference cohort ($p=.318$).

\par\medskip\noindent
\begin{minipage}{\columnwidth}
\small
\captionsetup{hypcap=false}
\captionof{table}{Research-choice differences across professional cohorts.}
\label{tab:research-priority-audit}
\begin{tabularx}{\columnwidth}{@{}Xrr@{}}
\toprule
\textbf{Comparison cohort} & \textbf{B1: $V$} & \textbf{B2: $V$} \\
\midrule
Wet-lab PhD / postdoc & 0.26 & 0.28 \\
Computational biologist & 0.25 & 0.27 \\
Molecular geneticist & 0.46 & 0.45 \\
\bottomrule
\end{tabularx}
\par\smallskip{\footnotesize\textit{Note.} Each comparison is against the senior tumor-immunology PI/research-scientist cohort, with 250 personas per group. B1 and B2 denote the primary and secondary research choices. Entries are Cram\'{e}r's $V$ for categorical response distributions; all six uncorrected $p$-values are below .001.}
\end{minipage}
\par\medskip

\paragraph{Insights.}
The simulated cohorts differed in research prioritization while showing no observed difference in the comprehension measure. This separates variation in scientific priorities from differences on that measured check. The strongest separation involved molecular geneticists, motivating discipline-specific analysis when designing expert consultation and subsequent research-planning studies.


\subsection{C5: Remote-Service Marketplace}\label{app:remote-service}

\paragraph{Summary.} A remote-service marketplace asked which inspection services to prioritize, whether customer and agent payment expectations overlapped, and what prevented participation. Because simulated price estimates had no behavioural baseline to check against, the entire study was run twice, under \texttt{gpt-5.6-sol} and \texttt{gpt-5.6-luna}, on the identical cohort, so that model sensitivity could be measured rather than assumed.

\paragraph{Cohort.} Two cohorts comprised 1,000 prospective customers and 1,000 prospective agents, drawn from synthesized personas with four quota-controlled dimensions each. Customers were aged 25--54 and distributed across North America (45\%), Western Europe (35\%), and Eastern Europe (20\%), with upper-middle (45\%), middle (35\%), and high (20\%) incomes, and a quota on trust disposition. Agents were distributed evenly across ten regions (10\% each), aged 18--64, spanning low to upper-middle incomes, with a quota on risk tolerance. Realized marginals deviated from target by at most 0.1 percentage points for customers and 1.4 for agents. All remaining attributes varied freely.

\paragraph{Task and design.} Each persona evaluated five scenarios: apartment inspection, used-car inspection, supplier verification, purchase verification, and document pickup. Scenario assignment was within-subject, so every persona answered all five and scenarios are compared within rather than between respondents. Questions covered clarity, trust in the individual agent, trust in the platform holding funds, participation intent, and payment expectations; agents additionally assessed personal safety. No price appeared in any scenario description, and payment expectations were elicited afterwards on a ten-rung ladder from \$10 to \$500 with an explicit refusal option, so that declining to transact was distinct from naming a low price. The design yielded 5,000 scenario-level responses per model per side, and 20,000 in total.

\paragraph{Results: scenario ordering.}
Table~\ref{tab:remote-service-results} compares indicative prices across models. The pooled clearing point was \$85 under \texttt{gpt-5.6-sol} and \$70 under \texttt{gpt-5.6-luna}. Document pickup had the lowest estimated price in both models; the highest-priced scenario was used-car inspection for Sol and apartment inspection for Luna. Scenario-price ranks had a correlation of 0.7.

\par\medskip\noindent
\begin{minipage}{\columnwidth}
\small
\captionsetup{hypcap=false}
\captionof{table}{Indicative service prices and matched cross-model agreement.}
\label{tab:remote-service-results}
\begin{tabularx}{\columnwidth}{@{}Xrr@{}}
\toprule
\textbf{Scenario} & \textbf{Sol (\$)} & \textbf{Luna (\$)} \\
\midrule
Apartment inspection & 86 & 93 \\
Used-car inspection & 100 & 88 \\
Purchase verification & 90 & 73 \\
Supplier verification & 80 & 68 \\
Document pickup & 65 & 48 \\
\addlinespace \multicolumn{3}{@{}l}{\textit{Matched price-rung agreement across models}} \\
\textbf{Participant group} & \textbf{Matched $N$} & \textbf{Exact agreement} \\
Customers & 4,995 & 27.0\% \\
Agents & 4,896 & 31.5\% \\
\bottomrule
\end{tabularx}
\par\smallskip{\footnotesize\textit{Note.} Models are \texttt{gpt-5.6-sol} and \texttt{gpt-5.6-luna}. Prices compare stated customer ceilings and agent floors after a 10\% commission; they are not observed market-clearing prices. Matched $N$ counts persona--scenario cells, not unique people.}
\end{minipage}
\par\medskip

\paragraph{Results: barriers to participation.} Reported situational relevance diverged sharply. \texttt{gpt-5.6-sol} treated essentially the whole cohort as addressable, at 97.3\% to 100\% across scenarios, whereas \texttt{gpt-5.6-luna} placed relevance between 30.9\% for document pickup and 75.2\% for supplier verification. Conditioned on relevance, however, the two models agreed closely on what blocks adoption: distrust of the agent was named as the principal barrier by 92.1\% of customers under \texttt{gpt-5.6-sol} and 95.6\% under \texttt{gpt-5.6-luna}, while price was named by 0.04\% under both. Customers trusted the platform holding their funds more than the individual performing the work in both models, at 3.99 against 3.70 under \texttt{gpt-5.6-sol} and 4.70 against 4.08 under \texttt{gpt-5.6-luna} on a five-point scale. Agents expressed greater confidence in being paid than in their personal safety, also under both models, at 4.01 against 3.15 and 4.74 against 3.05.

\paragraph{Results: model sensitivity.} Because both models answered for the same personas in the same scenarios, agreement can be measured within each persona-scenario cell rather than only between aggregate distributions. Across 4,995 matched customer cells the two models selected the same price rung 27.0\% of the time and landed within one rung 71.6\% of the time, with a Spearman correlation of 0.41 and a mean gap of \$25.40. The agent side agreed on 31.5\% of 4,896 cells, 82.6\% within one rung, with a correlation of 0.56. The matched comparisons reveal substantial sensitivity to model choice beyond differences visible in aggregate summaries.

\paragraph{Insights.} The findings distinguish situational relevance from enthusiasm, and financial feasibility from interpersonal trust. Apartment inspection emerged as a candidate for pilot testing, alongside agent verification, accountability, and safety procedures. The results also delimit what simulation can settle. The models identify the same principal participation barrier and broadly similar price rankings, supporting pilot design; absolute prices and addressable-market estimates remain model-dependent. The latter depends largely on how many people a model judges could ever be in the situation, which varied by a factor of two. Prices represent simulated stated preferences rather than validated demand or market-clearing prices, and the relevance estimate is the quantity most worth establishing against a real audience before any figure here is scaled to a population.

\subsection{C6: Professional-Network Signup}\label{app:professional-network}

\paragraph{Summary.}
This study examined whether a professional-networking website appeared relevant and trustworthy enough for prospective users to begin registration.

\paragraph{Cohort.}
The cohort comprised 1,000 simulated professionals aged 25--74 across ten regions, diverse roles, and industries. Profiles varied in technical confidence, privacy priorities, trust, identity-verification experience, and exposure to scams or bots.

\paragraph{Task and design.}
Using GPT-5.6-Luna, personas explored public pages and chose to start signup, defer, or reject. Visits were limited to six pages and permitted inspection of the signup entrance, but prohibited account creation or submission of personal information.

\paragraph{Results.}
All 1,000 trials completed. Table~\ref{tab:professional-network-results} shows that reaching and understanding the signup entrance rarely translated into willingness to register. Mean product clarity was 7.64/10, trust was 5.06/10, and signup intent was 2.24/5.

\par\medskip\noindent
\begin{minipage}{\columnwidth}
\small
\captionsetup{hypcap=false}
\captionof{table}{Website exploration and stated signup outcomes.}
\label{tab:professional-network-results}
\begin{tabularx}{\columnwidth}{@{}Xrr@{}}
\toprule
\textbf{Outcome} & \textbf{Count / valid $N$} & \textbf{Rate} \\
\midrule
Reached signup entrance & 976/1,000 & 97.6\% \\
Would begin registration & 9/1,000 & 0.9\% \\
Deferred registration & 806/1,000 & 80.6\% \\
Rejected registration & 185/1,000 & 18.5\% \\
Next step reported clear & 975/1,000 & 97.5\% \\
Trust as primary decision basis & 985/1,000 & 98.5\% \\
\bottomrule
\end{tabularx}
\par\smallskip{\footnotesize\textit{Note.} Counts are descriptive. Reaching the entrance is an exploration action; willingness to register is a stated choice. Neither measures completed onboarding.}
\end{minipage}
\par\medskip

\paragraph{Insights.}
Knowing how to register did not imply willingness to do so. Generated explanations associated hesitation with unclear verification procedures, document retention, and data handling despite relevant networking features. The findings identify pre-registration disclosures as candidates for revision. Reaching the entrance measures exploration, while choosing signup records stated willingness; neither establishes completed onboarding or human conversion.

\subsection{C7: Workplace Survey Responses}\label{app:workplace-assessment}

\paragraph{Summary.}
This study examined how simulated workers answered a workplace questionnaire, how team contexts changed responses, and whether model differences exceeded repeat variability.

\paragraph{Cohort.}
The cohort comprised 114 synthetic workers in eight teams across mining, manufacturing, transport, and other industrial services. Team sizes ranged from 5 to 38.

\paragraph{Task and design.}
Three linked experiments reused profiles and item order. GPT-4.1-mini answered an original 24-item questionnaire under mixed, supportive, and strained workplace contexts. GPT-4o-mini evaluated the baseline, while a 32-person panel completed three baseline calls per model. All 584 scheduled questionnaires were schema-valid.

\paragraph{Results.}
The favorable-coded baseline mean was 2.88/5; 64.8\% of raw answers were neutral. All 114 personas gave more favorable responses under the supportive context. Table~\ref{tab:workplace-results} reports this contrast and separates model differences from repeat variability. Baseline exact item agreement between models was 34.1\%, with a mean absolute difference of 0.769 points. AI-assisted inspection identified five clear value--explanation contradictions among 192 sampled answers.

\par\medskip\noindent
\begin{minipage}{\columnwidth}
\small
\captionsetup{hypcap=false}
\captionof{table}{Workplace-context effects and repeat-panel variability.}
\label{tab:workplace-results}
\begin{tabularx}{\columnwidth}{@{}Xrl@{}}
\toprule
\textbf{Comparison} & \textbf{Estimate} & \textbf{95\% interval} \\
\midrule
Supportive minus strained & $+3.85$ & [3.84, 3.87] \\
\addlinespace \multicolumn{3}{@{}l}{\textit{Mean absolute item differences in the repeated panel}} \\
Within GPT-4.1-mini & 0.192 & [0.181, 0.203] \\
Within GPT-4o-mini & 0.469 & [0.435, 0.509] \\
Between models & 0.754 & [0.720, 0.791] \\
Between minus mean within-model & 0.423 & [0.393, 0.457] \\
\bottomrule
\end{tabularx}
\par\smallskip{\footnotesize\textit{Note.} The context contrast uses 114 paired personas; the repeated panel uses 32 personas with three calls per model. Intervals use 10,000 complete-team resamples, preserving within-person responses. Effects are in questionnaire points; no significance labels are used.}
\end{minipage}
\par\medskip

\paragraph{Insights.}
Responses were strongly context-sensitive and model-dependent. Because six workplace conditions changed jointly, the context contrast does not isolate a single practice. Schema validity also did not ensure semantic consistency. The findings motivate explanation audits and repeat measurements before interpreting questionnaire summaries; the intervals describe synthetic teams, without human workforce benchmarking.


\subsection{C8: Industrial Website Engagement}\label{app:industrial-website}

\paragraph{Summary.}
An industrial software provider asked whether its Swedish homepage and four product pages helped prospective customers understand the offering and choose an appropriate next step.

\paragraph{Cohort.}
Separate page cohorts comprised 1,000 homepage visitors, 300 rental-system visitors, and 200 visitors each for equipment, fleet, and driving-log products. The partner's 90-day analytics informed the mix of prospective customers, existing customers, and other visitors. All 1,900 visits completed; 1,894 passed automated checks, including 1,088 prospect responses.

\paragraph{Task and design.}
Using Claude Opus 4.8, personas evaluated the first screen in a live browser, chose whether to continue, and inspected a next-step destination without entering personal information. They separately stated willingness to provide an email for a guide, receive a callback, attend a demo, or pilot the product on their own data. Each page was a one-cohort, one-stimulus study; browsing logs were retained alongside stated responses.

\paragraph{Results.}
Among prospects, 1,076/1,088 continued reading. Table~\ref{tab:industrial-website-results} shows that continued interest did not imply readiness for sales contact, and no prospect accepted a pilot. Rental-system demo willingness exceeded callback willingness, indicating that the commitment items were not a sequential conversion funnel. On fleet and driving-log pages, ``try for free'' led to sales forms; destination expectation-match ratings were 2.86/5 and 2.28/5. The rental demo and equipment savings calculator scored 4.97/5 and 4.96/5, respectively.

\par\medskip\noindent
\begin{minipage}{\columnwidth}
\small
\captionsetup{hypcap=false}
\captionof{table}{Stated commitments among prospective website visitors.}
\label{tab:industrial-website-results}
\begin{tabularx}{\columnwidth}{@{}Xrrrr@{}}
\toprule
\textbf{Page} & \textbf{$N$} & \textbf{Email} & \textbf{Call} & \textbf{Demo} \\
\midrule
Homepage & 499 & 377 & 50 & 0 \\
Rental system & 143 & 87 & 27 & 50 \\
Equipment & 150 & 52 & 4 & 5 \\
Fleet & 146 & 100 & 28 & 20 \\
Driving log & 150 & 104 & 3 & 1 \\
\bottomrule
\end{tabularx}
\par\smallskip{\footnotesize\textit{Note.} $N$ is the valid prospect-response denominator; entries count affirmative stated intentions. Pilot acceptance was zero on every page. Items were answered separately and do not measure completed conversions. Pages used separate cohorts; cross-page comparisons are descriptive.}
\end{minipage}
\par\medskip

\paragraph{Insights.}
The study distinguishes willingness to explore from willingness to engage with sales. Requests for customer references, reported by 86--100\% of prospects across pages, suggest that clearer proof and accurately labeled next steps merit testing before increasing commitment. These findings identify revisions for a subsequent pilot rather than establish human conversion rates.


\subsection{C9: Engagement-Letter Evaluation}\label{app:engagement-letter}

\paragraph{Summary.}
An AI-enablement partner supplied an illustrative accounting-service engagement letter and asked whether revised payment and liability terms would improve clients' willingness to proceed.

\paragraph{Cohort.}
The frozen cohort contained 400 personas for a scenario involving Indian businesses registered for goods and services tax (GST) renewing an accounting-firm relationship. Four economic-motivation categories crossed with four trust categories formed a balanced reference grid, with 25 personas per cell. GST status and the existing-client relationship were scenario constants, rather than verified individual attributes; the grid did not estimate client-population proportions.

\paragraph{Task and design.}
The same personas evaluated four letters in isolated trials. Payment schedules varied from 40/30/30 to 30/35/35, and liability caps from service-specific fees to total annual professional fees; the annual fee remained INR~235,000. GPT-5.6-Sol and Claude Opus 5 each completed 1,600 valid trials. The primary outcome combined signing as-is and requesting minor changes. The combined-versus-baseline contrast used exact paired McNemar inference; 19 secondary factorial contrasts used Benjamini--Hochberg correction.

\paragraph{Results.}
Both models showed positive primary contrasts (Table~\ref{tab:engagement-letter-results}; both exact $p<1.5\times10^{-11}$). Favorability increased by 0.1375 points for GPT and 0.0175 for Opus on a five-point scale. Neither met the working success rules, including 70\% acceptable intent and a 0.30-point favorability gain. In the combined condition, the 100-person hostile-trust group had no acceptable responses in either model.

\par\medskip\noindent
\begin{minipage}{\columnwidth}
\small
\captionsetup{hypcap=false}
\captionof{table}{Engagement-letter acceptance across simulation models.}
\label{tab:engagement-letter-results}
\begin{tabularx}{\columnwidth}{@{}XrrX@{}}
\toprule
\textbf{Model} & \textbf{Baseline} & \textbf{Combined} & \textbf{Lift (95\% CI)} \\
\midrule
GPT-5.6-Sol & 0.25\% & 28.75\% & $+28.50$\newline [24.06, 32.94] \\
Claude Opus 5 & 11.50\% & 20.75\% & $+9.25$\newline [6.40, 12.10] \\
\bottomrule
\end{tabularx}
\par\smallskip{\footnotesize\textit{Note.} Acceptance means signing as-is or requesting minor changes. Each model uses the same 400 personas per letter. Lifts and paired $t$-based 95\% intervals are in percentage points; significance uses exact McNemar tests. Intervals concern variation within the simulation design.}
\end{minipage}
\par\medskip

\paragraph{No-persona baseline.}
A separate GPT-5.6-Sol run completed 400 independent requests per letter without individual profiles, adding 1,600 valid responses. Acceptable intent was 0\% for baseline and payment-only, 46.25\% for liability-only, and 30.75\% for the combined revision. The combined-minus-liability-only acceptance contrast was +17.25 percentage points with personas and $-15.50$ without them. Their directly estimated difference was +32.75 percentage points (95\% CI [25.00, 40.50]).

\paragraph{Insights.}
The models agreed that joint revision improved acceptance but left the offer below the working threshold, despite differing in absolute response levels. This supports further testing of unresolved terms and low-trust audiences. The findings concern illustrative terms and simulated stated intent.


\subsection{C10: Healthcare Access Barriers}\label{app:healthcare-access}

\paragraph{Summary.}
This study examined which barriers a healthcare service should prioritize in India, including care-seeking and treatment management.

\paragraph{Cohort.}
We constructed 900 simulated personas across nine region--age groups: North, Central, and South India crossed with ages 18--30, 31--50, and 51--75, with 100 personas per group. Selected demographics were aligned with \href{https://censusindia.gov.in/census.website/en/data/population-finder}{Census 2011} data, and household resources with \href{https://microdata.gov.in/nada/index.php/catalog/237/related-materials}{Household Consumption Expenditure Survey (HCES) 2023--24} records. The cohort included 609 rural and 291 urban personas. Selected diagnosed conditions were calibrated using the \href{https://nhrdr.icmr.org.in/datasets/national-ncd-monitoring-survey-nnms-household-and-adult}{National Noncommunicable Disease Monitoring Survey (NNMS) 2017--18} and published Indian diabetes age tables.

\paragraph{Task and design.}
All personas completed the same English-language questionnaire covering eight areas, including care-seeking, medication-taking, and family influence. This was a response-characterization study with subgroup analyses.

\paragraph{Results.}
Table~\ref{tab:healthcare-access-results} shows common cost barriers, a pronounced rural--urban difference in distance barriers, and concentrated time constraints among ages 31--50. All 212 personas with fully scorable medication-adherence responses were classified as non-adherent under the study's scoring rules.

\par\medskip\noindent
\begin{minipage}{\columnwidth}
\small
\captionsetup{hypcap=false}
\captionof{table}{Healthcare-access barriers and subgroup patterns.}
\label{tab:healthcare-access-results}
\begin{tabularx}{\columnwidth}{@{}Xrr@{}}
\toprule
\textbf{Outcome / group} & \textbf{Count / valid $N$} & \textbf{Rate} \\
\midrule
Cost barrier: all personas & 780/900 & 86.7\% \\
Time constraints: all personas & 691/900 & 76.8\% \\
Distance barrier: rural & 527/609 & 86.5\% \\
Distance barrier: urban & 8/291 & 2.7\% \\
Time constraints: ages 31--50 & 277/300 & 92.3\% \\
Non-adherent: fully scorable responses & 212/212 & 100.0\% \\
\bottomrule
\end{tabularx}
\par\smallskip{\footnotesize\textit{Note.} Each row uses its eligible, valid-response denominator. Barriers permit multiple selections. Adherence classification applies only to fully scorable responses under the study scoring rules. These are descriptive simulated results.}
\end{minipage}
\par\medskip

\paragraph{Insights.}
The findings suggest three service priorities for further investigation: reducing financial barriers to care, improving transport accessibility for rural users, and accommodating work, household, and caregiving responsibilities when designing care services. Cost barriers were common across the cohort, distance barriers were concentrated among rural personas, and competing demands on time were most prominent among those aged 31--50. Uniform adherence classifications warrant examination of the instrument and simulation setup. These are findings from a reference-data-informed simulated cohort, without comparable human responses for behavioral benchmarking.

\subsection{C11: Home Assistant}\label{app:home-assistant}

\paragraph{Summary.}
This study examined whether a home assistant supports users with different comfort and device-runtime preferences, and whether those outcomes depend on users volunteering relevant details or waiting to be asked.

\paragraph{Cohort.}
The study used 50 base personas filtered for adulthood, North American location, native US culture, and California residence. Each base persona had four matched versions crossing comfort-first versus shorter-air-conditioning-runtime priorities with proactive versus on-request disclosure. Other persona attributes were held fixed across versions; this is a controlled reference cohort, not an estimate of population proportions.

\paragraph{Task and design.}
Users interacted with an assistant controlling devices such as thermostats, fans, and blinds. Two tasks introduced a complication: the arrival time changed while preparing the home, or the bedroom heat load increased while maintaining comfort until a set time. Each of GPT-5.6-Sol, Claude Opus 5, and Gemini 3.7 Flash simulated all four persona versions on both tasks, giving 400 scheduled trials per model. A programmatic verifier separately recorded goal completion, constraints and plan satisfaction, preference satisfaction, and scope compliance. The reported outcome is the binary preference check, not overall task success. For each base persona, the comfort-first minus shorter-runtime contrast averages over both disclosure conditions. Opus and Gemini interacted with DashScope's DeepSeek-V4.1-Flash; GPT interacted with DeepSeek's deepseek-flash endpoint, so cross-configuration comparisons also reflect an assistant-endpoint difference.

\paragraph{Results.}
Preference satisfaction was higher for comfort-first personas under all three simulation models (Table~\ref{tab:home-assistant-results}). The difference was 37.0--45.0 percentage points before arrival and 53.0--81.0 points for bedroom comfort. The Opus before-arrival comparison retains 49 complete four-condition personas because one trial lacks final component scores; all other comparisons retain 50. Missing component scores are not filled with zero. An invalid-action trial remains a failure in the source's overall-success accounting, which is distinct from this component analysis.

\par\medskip\noindent
\begin{minipage}{\columnwidth}
\small
\captionsetup{hypcap=false}
\captionof{table}{Home-assistant preference-satisfaction contrasts.}
\label{tab:home-assistant-results}
\begin{tabularx}{\columnwidth}{@{}XrX@{}}
\toprule
\textbf{Model} & \textbf{Difference} & \textbf{95\% CI} \\
\midrule
\multicolumn{3}{@{}l}{\textit{Prepare home before arrival}} \\
GPT-5.6-Sol & $+37.0$ & [25.0, 50.0] \\
Claude Opus 5 & $+37.8$ & [27.6, 48.0] \\
Gemini 3.7 Flash & $+45.0$ & [34.0, 56.0] \\
\midrule
\multicolumn{3}{@{}l}{\textit{Maintain bedroom comfort}} \\
GPT-5.6-Sol & $+81.0$ & [73.0, 88.0] \\
Claude Opus 5 & $+53.0$ & [42.0, 64.0] \\
Gemini 3.7 Flash & $+72.0$ & [63.0, 80.0] \\
\bottomrule
\end{tabularx}
\par\smallskip{\footnotesize\textit{Note.} Comfort-first minus shorter-runtime preference satisfaction, in percentage points, averaged over disclosure conditions. Intervals use 10,000 paired-persona bootstrap draws. $N=50$ complete personas except Opus before arrival ($N=49$). The scoring criterion follows each persona's priority; Appendix~\ref{app:stats} describes the pairing.}
\end{minipage}
\par\medskip

\paragraph{Insights.}
The recurring priority gap identifies uneven support for users' requirements: satisfying a comfort preference does not establish that the assistant can also minimize device runtime. Since each priority has its own preference criterion, the contrast is not a performance gain under one unchanged requirement. It motivates testing whether explicit trade-off elicitation and runtime-aware planning improve service for runtime-first users. This is a design hypothesis from simulated interactions, not evidence of measured energy savings or real-user satisfaction.

\subsection{C12: Car Assistant}\label{app:car-assistant}

\paragraph{Summary.}
This study examined whether an in-car assistant preserves travel preferences when users share them proactively versus only when asked, including situations where navigation plans must change.

\paragraph{Cohort.}
Fifty base personas used the same adulthood, North America, native US culture, and California-residence filters as C11. Four matched versions crossed fastest-arrival versus toll-avoidance priorities with proactive versus on-request disclosure. Comparisons retained the base persona and travel priority while changing disclosure behavior.

\paragraph{Task and design.}
Two navigation tasks were evaluated: adding a breakfast stop within opening hours under a tighter arrival deadline, and canceling an intermediate stop while keeping the arrival deadline when the first navigation update failed. Each of the three user-simulation models evaluated 50 personas in four conditions on both tasks, giving 400 scheduled trials per model. The assistant endpoints and programmatic component checks followed the setup described for C11. The main-text comparison uses preference satisfaction for the cancellation task; each contrast compares 50 matched proactive/on-request pairs within a fixed travel priority.

\paragraph{Results.}
In the cancellation task, proactive disclosure increased preference satisfaction under both priorities for every simulation model (Table~\ref{tab:car-assistant-results}). Differences ranged from 18.0 to 44.0 percentage points. In the breakfast-stop task, direction depended on the simulation model: under toll avoidance, the proactive-minus-on-request differences were $+4$, $-36$, and $+14$ points for GPT, Opus, and Gemini, respectively. That task therefore does not support the same cross-model directional claim.

\par\medskip\noindent
\begin{minipage}{\columnwidth}
\small
\captionsetup{hypcap=false}
\captionof{table}{Car-assistant preference-satisfaction contrasts.}
\label{tab:car-assistant-results}
\begin{tabularx}{\columnwidth}{@{}XrX@{}}
\toprule
\textbf{Model} & \textbf{Difference} & \textbf{95\% CI} \\
\midrule
\multicolumn{3}{@{}l}{\textit{Cancel stop: fastest-arrival priority}} \\
GPT-5.6-Sol & $+18.0$ & [6.0, 30.0] \\
Claude Opus 5 & $+44.0$ & [30.0, 58.0] \\
Gemini 3.7 Flash & $+36.0$ & [20.0, 52.0] \\
\midrule
\multicolumn{3}{@{}l}{\textit{Cancel stop: toll-avoidance priority}} \\
GPT-5.6-Sol & $+30.0$ & [16.0, 44.0] \\
Claude Opus 5 & $+44.0$ & [28.0, 60.0] \\
Gemini 3.7 Flash & $+42.0$ & [28.0, 56.0] \\
\bottomrule
\end{tabularx}
\par\smallskip{\footnotesize\textit{Note.} Proactive minus on-request disclosure, in percentage points, for canceling an intermediate stop while retaining the arrival deadline. Each row uses 50 matched persona pairs with travel priority held fixed. Intervals use 10,000 paired-persona bootstrap draws (Appendix~\ref{app:stats}).}
\end{minipage}
\par\medskip

\paragraph{Insights.}
The cancellation task exposes a service gap for users who wait to be asked: preserving their travel preference requires the assistant to elicit information rather than rely on unsolicited disclosure. The consistent positive contrast motivates testing preference clarification during navigation changes and recovery from failed updates. The breakfast-stop result limits the generalization: disclosure effects depend on the task and simulation configuration. These findings concern simulated navigation and preference checks, not validated driving safety or real-world travel outcomes.

\section{Statistical Reporting Protocol}\label{app:stats}
The protocol combines a common structure for trial-level evidence with analyses selected for each study's design. Descriptive evaluations report response distributions; comparison studies additionally report contrasts and the procedures used to assess them.

\paragraph{Outcome records and denominators.}
Each trial produces structured numerical, categorical, or textual outcomes. Reports retain measurement scales and category definitions, summarize numerical outcomes and categorical counts, and preserve the valid-response denominator for each metric. Missing responses are distinguished from recorded choices. Explanations are linked to the outcomes they describe, and aggregate results remain traceable to individual trial records.

\paragraph{Comparisons and uncertainty.}
The numerical comparison module matches responses by persona or pairing identifier. Matched comparisons use paired $t$-tests; unmatched comparisons use Welch's $t$-tests. Outputs include mean differences, standardized effects, and $t$-based confidence intervals, with 95\% confidence as the default. Categorical outcomes retain counts and proportions; the generic module also reports chi-square comparisons and Cram\'{e}r's $V$. These categorical tests assume independent arm observations and do not account for shared-persona pairing, so their inferential interpretation requires verification for repeated-person designs. The engagement-letter study uses exact paired McNemar inference for its binary primary outcome. Case-specific analyses accommodate other dependencies: the workplace study uses 10,000 whole-team resamples, preserving within-person responses, to obtain percentile intervals. The analysis or resampling unit is documented alongside the result.

\paragraph{Direct persona-condition comparisons.}
We use 10,000 bootstrap draws (seed 20261005) to obtain exploratory pointwise percentile 95\% intervals for differences between persona and no-persona conditions. In C1, complete persona pairs are resampled across framing arms; neutral responses are resampled independently by arm, and the two conditions independently. In C9, paired binary acceptance outcomes are resampled using the joint counts recovered from published paired sufficient statistics; no-persona responses are resampled independently by letter. Each reported difference is persona minus no-persona. The direct differences reported for C1 and C9 are 0.217 trust points in C1 (95\% CI [0.183, 0.251]), and 32.75 percentage points in the combined-minus-liability acceptance contrast in C9 ([25.00, 40.50]).

\paragraph{Intervals for plotted estimates.}
Error bars in Figures~\ref{fig:model-persona-comparisons} and~\ref{fig:persona-information-comparisons} use 10,000 pointwise percentile bootstrap draws (seed 20261005). For home assistance, each resampled base persona retains all four priority-by-disclosure conditions; the comfort-first minus short-runtime contrast is averaged across its two disclosure conditions. The Opus before-arrival comparison uses 49 complete four-condition personas because one component score is missing; all other home comparisons use 50. Car contrasts resample 50 personas with their two disclosure outcomes kept together within each priority, reporting proactive minus on-request preference satisfaction. Both cases use the recorded programmatic preference check. Clinical intervals resample categorical responses within each of four professional groups, retaining 200 responses per group. Engagement-letter differences resample 400 paired persona outcomes: Opus and Gemini pairs are matched by persona ID, and GPT uses the uniquely recovered joint counts from published sufficient statistics. The GPT no-persona arms are resampled independently, with 400 responses per letter. Home, car, and engagement intervals apply to the plotted differences; clinical intervals apply to Cram\'{e}r's $V$.

\paragraph{Multiple comparisons and subgroup analysis.}
Benjamini--Hochberg correction is available for the family of tested arm--metric comparisons and is optional; the default is uncorrected inference. Reports identify the correction used. Original music-wellbeing reports apply correction to overall comparisons; the local-data contrasts and new bootstrap intervals reported here are pointwise and unadjusted. The engagement-letter study applies correction to 19 secondary factorial contrasts, with a single prespecified primary contrast; its new direct persona-condition interval is also pointwise and unadjusted. Generic subgroup significance labels use raw $p$-values and are treated as exploratory unless a separate correction is documented. Outcome-linked explanations support interpretation without constituting an independent statistical test.

\paragraph{Scope of the intervals.}
The intervals describe variation within the specified simulation design, conditional on the model, cohort construction, and task. With one response per persona and condition, observed variation combines differences across personas with stochastic generation. Repeated trials with fixed personas further assess generation variability, as in the workplace study. Neither these intervals nor agreement across models quantifies discrepancies from real-user behavior; that assessment requires matched human reference data.

\end{document}